\documentclass{article}
\usepackage{PRIMEarxiv}
\usepackage{graphicx}
\usepackage{hyperref}
\usepackage{url}
\usepackage{cite}
\usepackage{amsmath}
\usepackage{gensymb}
\usepackage{float}

\title{From drift to adaptation to the failed ml model: Transfer Learning in Industrial MLOps}

\author{
Waqar Muhammad Ashraf \\
University College London \\
The Alan Turing Institute \\
London, United Kingdom
\And
Talha Ansar \\
Univeristy of Engineering and Technology \\
Lahore, Pakistan
\And
Fahad Ahmed \\
Sahiwal Coal Fired Power Plant \\
Sahiwal, Pakistan
\And
Jawad Hussain \\
Sahiwal Coal Fired Power Plant \\
Sahiwal, Pakistan
\And
Muhammad Mujtaba Abbas \\
Univeristy of Engineering and Technology \\
Lahore, Pakistan
\And
Vivek Dua \thanks{Corresponding author: v.dua@ucl.ac.uk} \\
University College London \\
London, United Kingdom
}

\begin{document}
\maketitle
\begin{abstract}
Model adaptation to production environment is critical for reliable Machine Learning Operations (MLOps), less attention is paid to developing systematic framework for updating the ML models when they fail under data drift. This paper compares the transfer learning enabled model update strategies including ensemble transfer learning (ETL), all-layers transfer learning (ALTL), and last-layer transfer learning (LLTL) for updating the failed feedforward artificial neural network (ANN) model. The flue gas differential pressure across the air preheater unit installed in a 660 MW thermal power plant is analyzed as a case study since it mimics the batch processes due to load cycling in the power plant. Updating the failed ANN model by three transfer learning techniques reveals that ETL provides relatively higher predictive accuracy for the batch size of 5 days than those of LLTL and ALTL. However, ALTL is found to be suitable for effective update of the model trained on large batch size (8 days). A mixed trend is observed for computational requirement (hyperparameter tuning and model training) of model update techniques for different batch sizes. These fundamental and empiric insights obtained from the batch process-based industrial case study can assist the MLOps practitioners in adapting the failed models to data drifts for the accurate monitoring of industrial processes.
\end{abstract}

\keywords{Model failure \and Data drift \and Human centered monitoring \and Industrial processes}

\section{Introduction}

The industrial operations have nonlinear characteristics, meaning that change in the process conditions over time and equipment changeover may result in changing patterns of correlations between the process variables. The “static” data-driven models trained on “fixed data-distributions” cannot handle non-stationary data streams of industrial processes and are not suitable for their deployment for long-term monitoring tasks. This is attributed to data drift (shift in covariance due to change in data distributions between the input variables of process) and/or concept drift (the underlying relationships between the input and output variables of process are changed) that drive the model failure.
The data-driven models are retrained on the new data streams in order to maintain their predictive accuracy for new data-distributions. The potential caveat with this approach is the “catastrophic forgetting”, meaning that the model parameters have been tuned to the new dataset and may not predict well the historical data the model was trained on. Initially, Geoffrey E. Hinton hypothesized that small changes in model parameters for their tuning to new data may collectively cancel out their effect on historical data \cite{Hinton1984}. However, McCloskey and Cohen \cite{McCloskey1989} and Ratcliff \cite{Ratcliff1990} showed that sequential training of simple neural networks on disjoint sets of data results in significant forgetting even when the models are updated on small amount of new data.
To address the potential issue of catastrophic forgetting, continual learning approach is used in updating the model parameters and it balances the trade-off between the plasticity (learn new dynamics) and stability (retaining historical information) through experience replay and regularization techniques during the model update mechanism. On the other hand, online learning retrains the model on a single or batch of data continuously once the model is productionised. Online learning offers high adaptability to the model to the incoming data stream or concept drift; however, it suffers from catastrophic forgetting. Just-In-Time-Learning is another model update paradigm where a local model is trained on the incoming data stream and the similar data queried from the historical data. Just-In-Time-Learning offers high accuracy to approximate nonlinear dynamics of system. However, it has higher computational and memory requirement and may not extrapolate when system enters in the new operating model where historical data is not available. To enable the high predictive accuracy and reducing the computational expense for the models in production, transfer learning approach is also implemented where parameters of trained model (source domain maintained on historical data) are transferred for adapting them with the “target” domain (drift in data) through minimal requirement of data.
The literature studies identify the importance of updating the model for their adaptability to the changing industrial operations. In \cite{Rebello2025}, the neural network parameters in all layers are tuned upon the detection of drift and resulting updated model was deployed for digital twin applications. In another study \cite{Metcalfe2025}, the researchers have allocated the epoch budget and bounds on learning rate for updating the model parameters completely. Similarly, transfer learning is also implemented for model adaptation to source domain \cite{Theisen2025}. However, it is noted that much of research attention is paid to update the model parameters, the effect of batch-size on updating the model parameters by transfer learning-based model update techniques is not studied systematically. More importantly, it remains largely unknown how transfer learning driven model update techniques including ensemble transfer learning (ETL), all-layers transfer learning (ALTL), and last-layer transfer learning (LLTL) for neural networks in combination with batch size and how it would affect the predictive accuracy of the model. These fundamental issues about the selection of model update techniques considering plasticity vs stability and the computational expense are not investigated systematically to obtain mechanistic insights about addressing these issues.
Motivated by the current research gaps, this paper investigates the effect of batch size and transfer learning driven model update techniques (ETL, ALTL, LLTL) to address the concept drift for updating the feedforward ANN model. The flue gas differential pressure (DP) across the air preheater installed in a 660 MW thermal power plant is considered as a case study. Flue gas DP is correlated with load cycling (ramp-up and ramp down) and is treated as a batch process which is observable from the data visualizations. This allows to transfer the findings from this case study to the dynamics of batch processes which are carried out in various chemical engineering applications. A systematic comparison is made to reveal the effect of batch size on the predictive accuracy of updated model by three techniques, changes in the weight space of the updated models as well as their computational requirement. Moreover, the evolution in the feature importance is also investigated by SHapley Additive Explanations (SHAP) when the models are updated to reveal their stability for industrial monitoring tasks. These fundamental and empiric insights obtained through the industrial case study can assist the MLOps practitioners in adapting the failed models to data drifts for the accurate monitoring of industrial processes.

\section{Method}
\begin{figure}[t]
  \centering
  \includegraphics[width=0.7\linewidth]{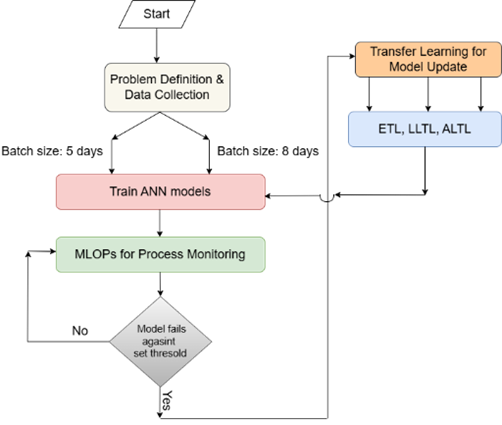}
  \caption{Schematic on transfer learning-based model update framework for process monitoring.}
  \label{fig:1}
\end{figure}

Figure~\ref{fig:1} shows the transfer learning enabled model update framework developed in this work to monitor the flue gas DP across the air preheater. Two months of data is collected separately for the month of January and April for the model development on the two batch sizes of 5 days and 8 days. Initially, a set of two batches (for 5 days and 8 days each) is used to train ANN models under hyperparameters optimization (learning rate, number of neurons in the hidden layer). The models are evaluated on coefficient of determination (R$_2$) and root-mean-squared-error (RMSE) on the test data (20\% of set). Later, the trained models are deployed as MLOPs to predict the Flue Gas DP for the rest of batch data. When daily average error (RMSE and MAE) remains above the twice of test data (observed during model training) for three times consecutively, model update mechanism is triggered. A set of the batch data where model fails and the consecutive previous batch data (buffer data for model retraining) is used to update the model by transfer learning driven approaches (ETL, LLTL and ALTL). Later, the models trained by the three approaches are put into MLOPs again to predict the rest of the collected data. This allows to compare the predictive accuracy of the model against the ground truth already available in the collected data. The stability of the models in terms of evolution of feature importance for the trained and updated models under different batches and model update approaches are investigated by SHAP framework. The feature evolution is affected by the process dynamics; however, they can be corrupted by the ineffective model update and mapping created in the model. The computational time consumed to train and update the models is also recorded to investigate the suitability of transfer learning enabled model update approaches for MLOPs.

\section{Results}
\subsection{Data Characteristics}

\begin{figure}[t]
  \centering
  \includegraphics[width=0.9\linewidth]{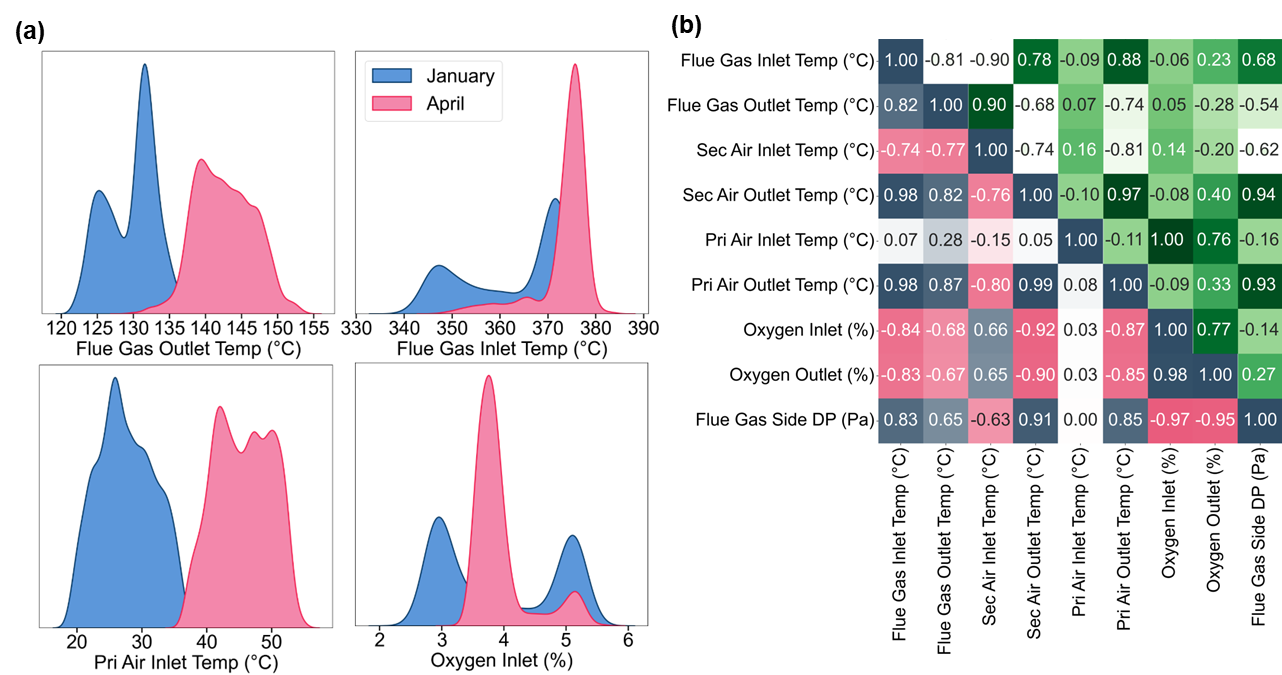}
  \caption{Characteristics of data collected for heat exchange across air preheater. (a) Distribution shift is observable in data collected for month of January and April. (b) Correlation structure between the operating variables is also changed for the two months (lower diagonal: January, upper diagonal: April).}
  \label{fig:2}
\end{figure}

The operating variables across the air preheater for modelling flue gas DP (Pa) are as follows: flue gas inlet temperature ($\degree$C), flue gas outlet temperature ($\degree$C), secondary air inlet temperature ($\degree$C), secondary air outlet temperature ($\degree$C), primary air inlet temperature ($\degree$C), primary air outlet temperature ($\degree$C), oxygen inlet (\%), oxygen outlet (\%). The historical operation data of the variables is collected at the averaging interval of 10 minutes and 8 minutes, respectively, for the month of January and April in order to explore the capacity and efficacy of models to handle different sampling frequency of data collection for the monitoring task. This allows us to mimic the real-life situation when sampling frequency of data collection is affected by data storage and maintenance costs and in turn investigate how the model responds to this transition yet ensuring predictive accuracy.
 Figure~\ref{fig:2}(a) presents the data-distribution profiles of a few operating variables for the months of January and April. The asymmetric data-distribution as well as data drift are visually observable for the variables and further quantified by population stability index (PSI) and Cramér-von Mises (CvM). PSI ranges from 1.89 to 23.15 and CvM statistics from 72.4 to 658.9 with p-values less than 0.001 confirm drift in the data observed in two datasets sampled at multi time-scale sampling. Primary air inlet temperature exhibits the strongest drift across both metrics (PSI=23.15, CvM=658.9) and flue gas DP (PSI=1.89) and oxygen outlet (CvM=72.4) show the relatively smaller drift, yet noticeable for potential model failure. The correlation between the pair of variables is computed for the two months separately (shown on Figure~\ref{fig:2}(b)) and reveals that correlation between the pair of variables is changed in the two datasets (concept drift is present in the data). This indicates different operating modes of operation are observed for the air preheater that is useful for the monitoring evaluation capacity of the model once deployed in production.
\clearpage

\subsection{Training of ANN model and update under two different batch sizes}
\begin{figure}[t]
  \centering
  \includegraphics[width=1\linewidth]{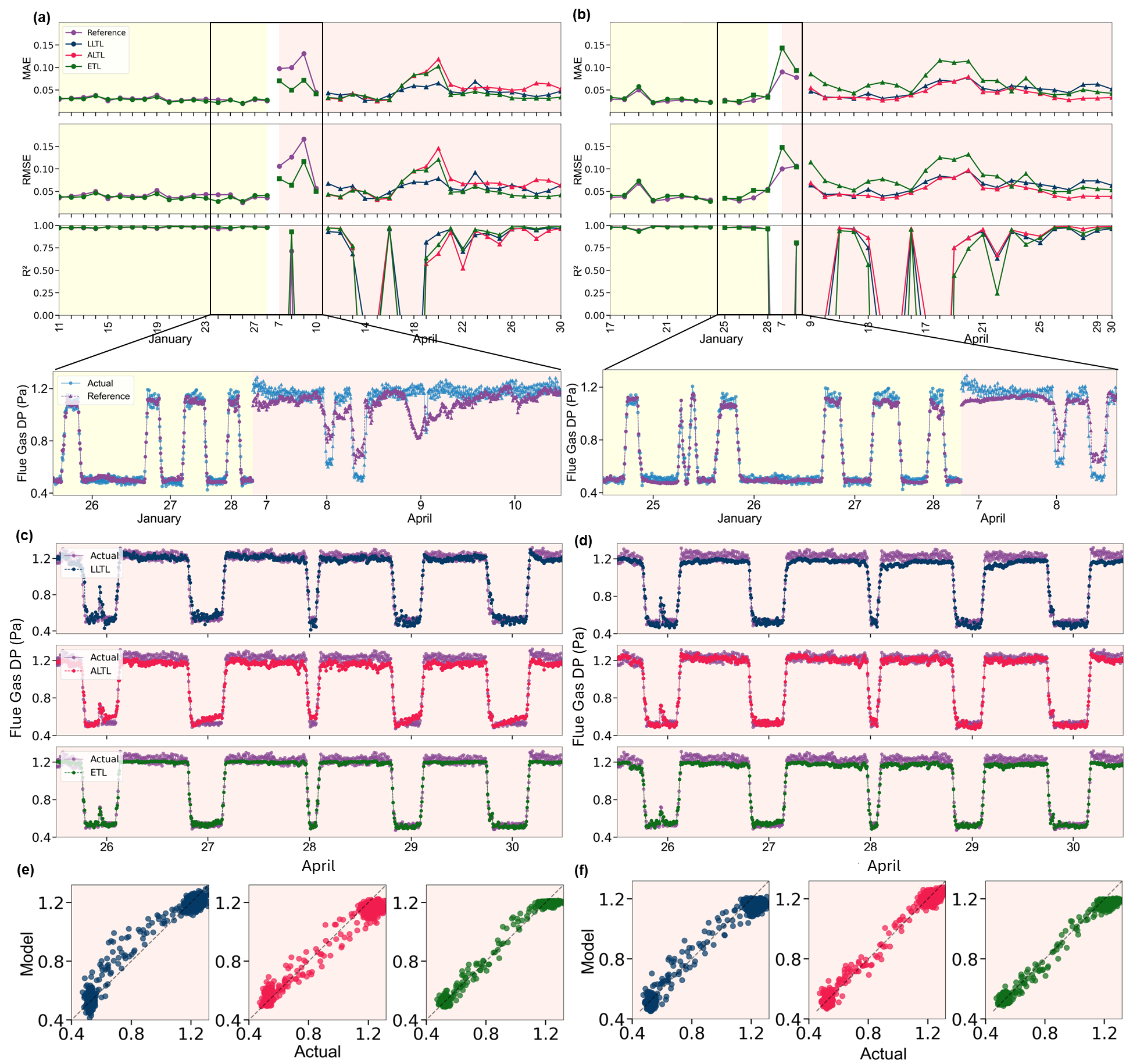}
  \caption{MLOps of ANN models trained on batch sizes of (a) 5 days and (b) 8 days to monitor flue gas DP across air preheater. The trained ANN model exhibited good monitoring performance in the month of January. However, they failed during 7-10 April and model update techniques are implemented. Actual and model-based responses are also mapped corresponding to the failure zone. Peformance of models updated by LLTL, ALTL, ETL techniques on remaining April data is shown in (c,d). (e,f) parity plot to compare actual and model-based response for three model update techniques in the remaining april data.}
  \label{fig:3}
\end{figure}

Feedforward ANN models are trained on a set of two batches of 5 days and 8 days separately, taken from the data of January. During the training, hyperparameters of the ANN model (hidden layer neurons, learning size, number of hidden layers) are optimized and models are evaluated on the test data (20\% of set).  The model trained on set of 5-day batches yielded a test R$_2$ of 0.9824 (RMSE 0.0381), while the model on set of 8-day batches achieved a test R$_2$ of 0.9807 (RMSE 0.0385). After the training phase, the models trained on batch size of 5 days and 8 days are deployed in production to monitor flue gas DP across the air preheater. The monitoring accuracy of the two models is depicted on Figure~\ref{fig:3}(a,b) respectively. It is noted that the ANN model trained on batch size of 5 days exhibited relatively superior performance in predicting the Flue Gas DP in the month of January than those of ANN model trained on batch size of 8 days.
Between April 7 and 10, the two models crossed the human-defined error threshold, i.e., crossing twice the test error limit successively for three times. The mapping of actual and model-based responses (reference) is shown in the bottom of Figure~\ref{fig:3}(a,b). The failure of models on different sampling frequency of data collection reveals that the trained models do not scale well when data sampling frequency changes with respect to those the model was trained on. The failure of model can also be attributed to process dynamics and data drift as depicted on Figure!\ref{fig:2}. The model failure triggers the model update procedure built on three techniques including ETL, ALTL, and LLTL and we carried it out for batch size of 5 days and 8 days separately (batch data where model fails and the consecutive previous batch). It is important to note here that during model update, only learning rate is optimized iteratively in the narrow range and without changing the model architecture. Later, the models trained on three model update techniques are put into MLOPs to monitor Flue Gas DP across the air preheater for the rest of data of April.
The mapping of actual and model-based responses obtained from models trained by ETL, ALTL and LLTL is depicted on Figure~\ref{fig:3}(c,d) for the batch size of 5 days and 8 days respectively. Similarly, parity plots are also constructed to directly compare the actual and model-based responses as shown on Figure 3(e,f). The RMSE and MAE for ETL are as follows: 0.0564, 0.0414 (5-day), 0.0652, 0.0541 (8-day); ALTL: 0.0674, 0.0563 (5-day), 0.0384, 0.0315(8-day); and LLTL 0.0400, 0.0321 (5-day), 0.0537, 0.0438 (8-day). The visualization of actual and model-based responses in conjunction with errors computed for the models trained on three update techniques allows us to inspect local predictive performance of the trained model once productionized in industrial settings. 
ETL approach seems to monitor flue gas DP with relatively more accuracy in comparison with ALTL and LLTL for the model trained on batch size of 5 days (Figure~\ref{fig:3}(a)). On the other hand, ALTL is found to have comparatively better predictive performance for the model trained on batch size of 8 days (Figure~\ref{fig:3}(b)). For small batch sizes, ensemble may average out the predictive responses and provide stable performance. However, when batch size increases, ALTL updates the model parameters across all layers and model is adapted with the recent data stream. This increases the risk of “forgetting” the operation history and model is poised to adapt with the recent operation dynamics.

\subsection{Insights about layer-wise weight space distribution after model update}
\begin{figure}[H]
  \centering
  \includegraphics[width=1\linewidth]{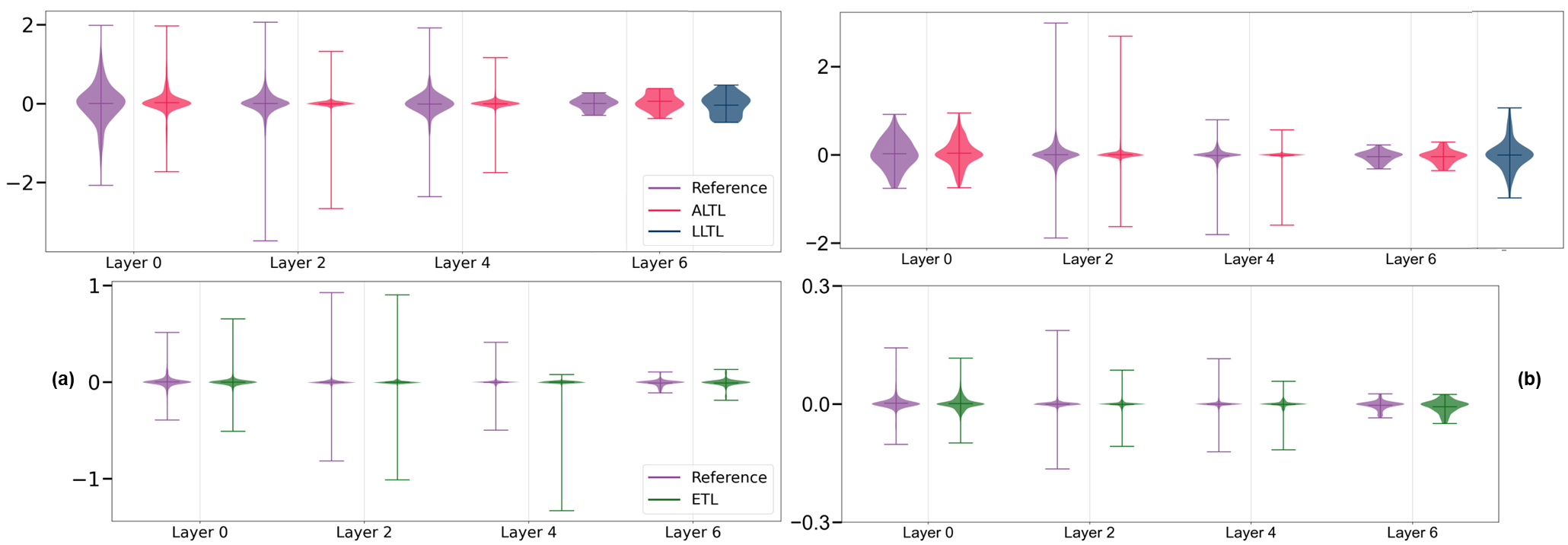}
  \caption{Weight space distribution changes with model update for the batch size of (a) 5 days, and (b) 8 days. In common, last-layer weight space is widened to adapt with data-drift.}
  \label{fig:4}
\end{figure}

The layer-wise weight space of the models updated with three techniques is retrieved and is visualised to obtain insights on how model parameters are tuned or varied after the model update phase corresponding to batch size of 5 days and 8 days. The weight-space distribution is visualized as violin plot for each layer of ANN and is depicted on Figure~\ref{fig:4}(a,b) corresponding to batch size of 5 days and 8 days respectively. In general, it is found that weight space distribution for last-layer of ANN gets widened for the models trained on batch sizes of 5 days and 8 days. This can be attributed to back-propagation of error signal starting from last layer of ANN that tunes the model parameters. For ALTL, the weight space of hidden layers seems to be compressed than those of reference distribution (pre-trained model space) that can explain the improved generalization capability of the model while adapting with drift observed in the data distribution collected for the month of April. However, ensembles approach seems to depict mixed results in widening or compressing the weight space that can be explained by a number of ensembles digesting the data based on the batch sizes.

\subsection{Evolution of feature importance}

\begin{figure}[H]
  \centering
  \includegraphics[width=0.9\linewidth]{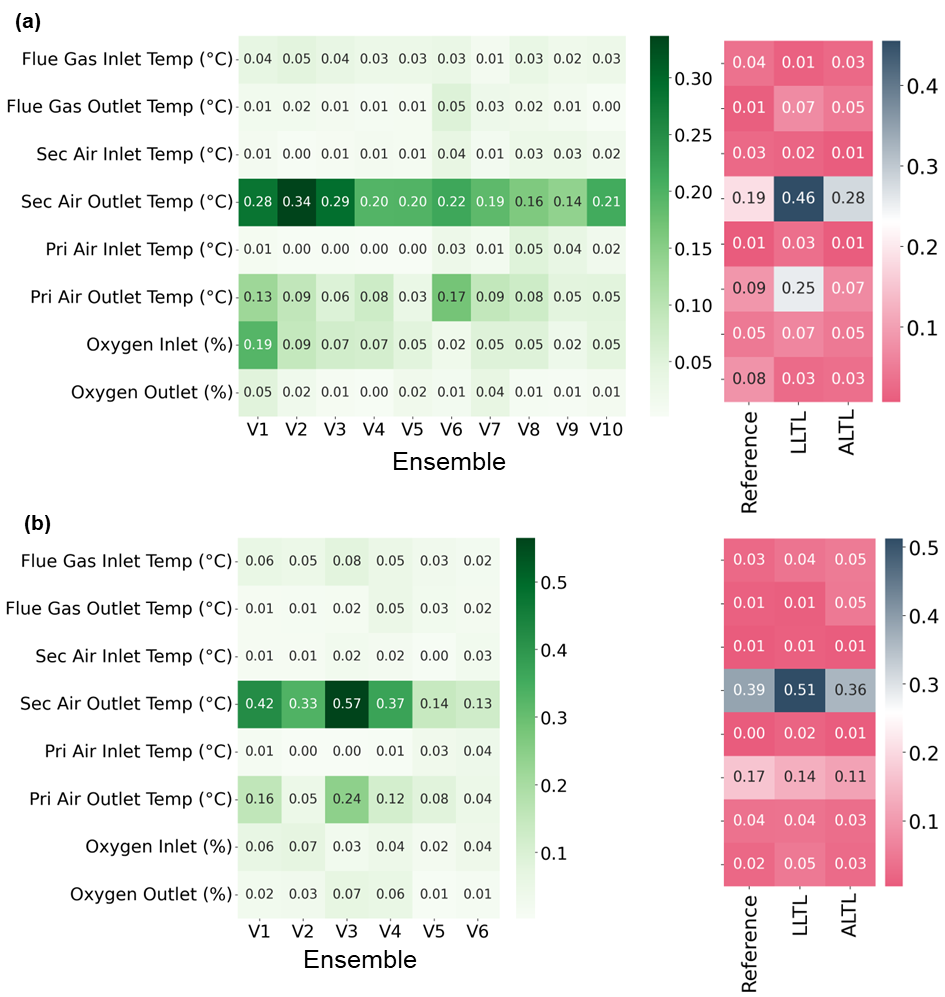}
  \caption{Evolution of SHAP-based feature importance with model update for ETL, LLTL and ALTL with respect to batch size of (a) 5 days and (b) 8 days. While Secondary Air Outlet Temperature (°C) is consistently the most dominant feature, but during periods of information scarcity, Primary Air Outlet Temperature (°C) exhibits a transient increase in importance to compensate for the lack of information of the new regime (V6 in (a) and V3 in (b)). Once sufficient data is accumulated, the model stabilizes, and Sec Air Outlet Temperature (°C) regains its primary significance.}
  \label{fig:5}
\end{figure}

The feature importance to predict flue gas DP is established by SHAP and is used to investigate the evolution in feature importance after the model update. The reference column represents the feature importance before the model update that serves as a baseline to compare the feature importance order after the model update. Secondary air outlet temperature is found to be the most significant feature affecting flue gas DP for models updated on batch size of 5 days and 8 days, respectively, as shown on Figure~\ref{fig:5}(a,b). After the model update, secondary air outlet temperature remains the most significant feature as established for models updated by LLTL, ALTL and ETL approaches. Multiple ETL columns explicitly track the sequential updates across successive batch intervals, where each column corresponds to a distinct version produced after an update and the resulting feature importance order. The consistent significant order of features, especially for most significant operating variable, by three model update techniques depicts the accuracy of feature importance analysis as well as stability of the trained models to capture the process dynamics. It also reveals that feature importance may remain the same even when concept and data drift are observed in the data (refer to Figure~\ref{fig:2}).
\clearpage

Although the magnitude of feature importance is varied for each model update approaches, the feature importance order remains nearly the same, especially for significant features. This may depict the stable tuning of model parameters, irrespective of batch-sizes, and is also evident as stable error profiles depicted on Figure~\ref{fig:3}(a,b).

\subsection{Computational Performance of model update techniques}
\begin{figure}[H]
  \centering
  \includegraphics[width=0.9\linewidth]{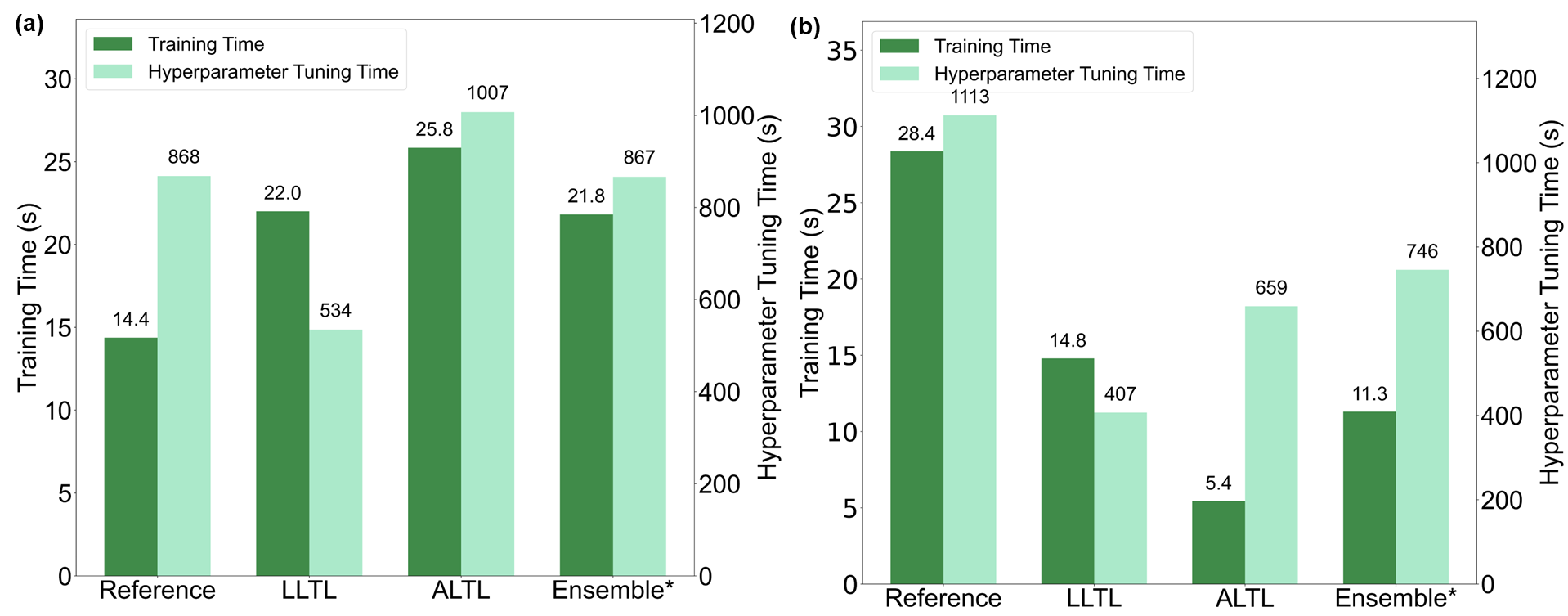}
  \caption{Comparing the computational time consumed for hyperparameter tuning and model training for batch size of (a) 5 days and (b) 8 days.}
  \label{fig:6}
\end{figure}

The computational time required to train a reference model as well as its update by three techniques for the batch size of 5 days and 8 eights is recorded and is presented on Figure~\ref{fig:6}(a,b). For a reference model, comparatively larger computational time is observed (hyperparameter tuning time: 1113 s, training time: 28.4 s) for model trained on batch size of 8 days than those of model trained on 5 days (hyperparameter tuning time: 868 s, training time: 14.4 s). It can be attributed to nonsmooth convergence to error minimum on the fewer batches. However, models trained on batch size of 5 days, in general, consumed large computational time for model update by LLTL, ALTL and ETL approaches than those of models trained on batch size of 8 eights. This observation may reflect that stochastic error convergence in the baseline model training produces the parametric values that adapt quickly on new data streams as compared to smoothly converged models. A mixed trend is observed in comparing the computational time used for updating the models for the same batch size. However, hyperparameter tuning time remains larger than those of training time irrespective of batch sizes.

\section{Conclusion}

We provide insights into the selection of model update techniques when feedforward ANN model fails in MLOps due to drift (data and concept). The effect of batch size (5 days or 8 eights) on the predictive accuracy of feed forward ANN model is investigated to monitor flue gas DP across air preheater (treated as batch process) installed in a 660 MW thermal power plant. We find that model failure is driven by data drift and ensemble approach implemented for batch size of 5 days (small batch size) provides superior monitoring performance than those of LLTL and ALTL techniques. This can be attributed to widening of weight space across the layers of ANN models in the ensemble which is otherwise compressed for LLTL and ALTL techniques. The comparison of computational time requirements (hyperparameter tuning and model training) for model update techniques reveals that ETL takes longer to update the model for the batch size of 5 days. However, ALTL, which depicts superior monitoring performance for the large batch size, takes the lowest time for updating the model once hyperparameters have been optimised. This aspect has wider implications for MLOPs practices since the models having stochastic error convergence profiles may quickly adapt with drift (data and concept) than the models which initially have smooth error convergence profiles. These fundamental insights obtained from model update for industrial batch processes highlight the role of batch size of data and the selection of model update techniques for industrial monitoring applications. These findings can also help troubleshoot the MLOps pipeline and dealing with failed ML model for adapting it to volatile industrial production environments.

\bibliographystyle{unsrt}
\bibliography{references}

\end{document}